\documentclass[conference]{IEEEtran}
\IEEEoverridecommandlockouts
\usepackage{amsmath,amssymb,amsfonts}
\usepackage{algorithmic}
\usepackage{graphicx}
\usepackage{textcomp}
\usepackage{xcolor}
\usepackage[noend]{algorithm2e}
\usepackage{amssymb}
\usepackage{amsthm}
\usepackage{breqn}
\usepackage{mathtools}
\usepackage{url}

\DeclarePairedDelimiter\floor{\lfloor}{\rfloor}

\newcommand{\rank}{\mathrm{rank}}
\newcommand{\minimize}[1]{\underset{#1}{\textrm{minimize}}}
\newcommand*{\Scale}[2][4]{\scalebox{#1}{$#2$}}%

\newtheorem{defi}{Definition}
\newtheorem{prop}{Proposition}
\newtheorem{obs}{Observation}

\newtheorem{ass}{Assumption}

\newtheorem{exam}{Example}

\newtheorem{theorem}{Theorem}
\usepackage{color, soul}

\def\BibTeX{{\rm B\kern-.05em{\sc i\kern-.025em b}\kern-.08em
    T\kern-.1667em\lower.7ex\hbox{E}\kern-.125emX}}
\begin{document}

\title{Improving Fairness for Data Valuation in Horizontal Federated Learning}

\author{
\IEEEauthorblockN{
Zhenan Fan\IEEEauthorrefmark{1},
Huang Fang\IEEEauthorrefmark{1}, 
Zirui Zhou\IEEEauthorrefmark{2},
Jian Pei\IEEEauthorrefmark{3},
Michael P. Friedlander\IEEEauthorrefmark{1},
Changxin Liu\IEEEauthorrefmark{4} and
Yong Zhang\IEEEauthorrefmark{2}}
\vspace{3mm}
\IEEEauthorblockA{
\IEEEauthorrefmark{1} University of British Columbia, \textit{\{zhenanf, hgfang, mpf\}@cs.ubc.ca}\\
\IEEEauthorrefmark{2} Huawei Technologies Canada Co., \textit{\{zirui.zhou,yong.zhang3\}@huawei.com}\\
\IEEEauthorrefmark{3} Simon Fraser University, \textit{jpei@cs.sfu.ca}\\
\IEEEauthorrefmark{4} KTH Royal Institute of Technology, \textit{changxin@kth.se}}
}

\maketitle

\begin{abstract}
Federated learning is an emerging decentralized machine learning scheme that allows multiple data owners to work collaboratively while ensuring data privacy. The success of federated learning depends largely on the participation of data owners. To sustain and encourage data owners' participation, it is crucial to fairly evaluate the quality of the data provided by the data owners as well as their contribution to the final model and reward them correspondingly. Federated Shapley value, recently proposed by Wang et al. [Federated Learning, 2020], is a measure for data value under the framework of federated learning that satisfies many desired properties for data valuation. However, there are still factors of potential unfairness in the design of federated Shapley value because two data owners with the same local data may not receive the same evaluation. We propose a new measure called completed federated Shapley value to improve the fairness of federated Shapley value. The design depends on completing a matrix consisting of all the possible contributions by different subsets of the data owners. It is shown under mild conditions that this matrix is approximately low-rank by leveraging concepts and tools from optimization. Both theoretical analysis and empirical evaluation verify that the proposed measure does improve fairness in many circumstances. 
\end{abstract}

\begin{IEEEkeywords}
contribution evaluation, fairness, federated learning
\end{IEEEkeywords}

\section{Introduction} \label{sec:1}

Building large and powerful machine learning models often needs large amounts of training data. In many industry-scale applications, training data is distributed in the form of silos, that is, training data is obtained and maintained by many data owners instead of being centralized at the place of a single owner or a data center. Because of industrial competition, privacy concerns, legal restrictions, and many other possible reasons, integrating or centralizing data from different sources faces enormous resistance and is often even infeasible. Federated learning~\cite{federated2016} is promising for training machine learning models on distributed data sources. It facilitates collaboration among a group of data owners (aka.~``clients'') and, at the same time, preserves their privacy. The central idea of federated learning is to periodically aggregate local models from clients to produce a more general and capable global model.

Federated learning can be divided into two classes: horizontal federated learning and vertical federated learning~\cite{yang2019federated}. Horizontal federated learning refers to the scenarios where data sets owned by different clients share the same feature space but differ in samples. Vertical federated learning refers to the scenarios where data sets owned by different clients share the same sample IDs, but have different features. In this work, we focus on the horizontal federated learning. 

The success of federated learning relies on active participation by motivated data owners. The motivation of data owners partially depends on whether the collaboration and rewarding in federated learning are fair.  Thus, it is essential to understand how to fairly and efficiently evaluate the contributions of data owners in federated learning~\cite{zhang2020hierarchically,song2019profit,wei2020efficient,wang2020principled}. In this paper, we focus on the fairness in contribution valuation.


Shapley value~\cite{shapley201617} is a classical measure originates from cooperative game theory to fairly assess contributions by participants in a coalition. 
The Shapley value of a participant is defined as the expectation of the marginal contribution of the participant over all possible subsets of the other participants. Shapley value is the unique measure that satisfies the four fundamental requirements of fairness proposed by Shapley~\cite{shapley201617}: balance, symmetry, zero element and additivity (see a brief review in Section~\ref{sec:fair}). Although Shapley value has many desirable properties, evaluating Shapley value in federated learning requires exhaustive retraining and evaluating the model on every subset of clients. The costs of communication and time may be prohibitive in practice~\cite{song2019profit}.

To make fair evaluation of data owners practical in federated learning, some variations inspired by Shapley value were proposed. For example, Wang \textit{et al.}~\cite{wang2020principled} recently proposed federated Shapley value (FedSV). The key idea is to compute the Shapley values for clients in each round of training and then report the summation over all the rounds as the final results. Compared with the classical Shapley value, FedSV does not require model retraining and retains some but not all of the desirable properties for fairness. 

However, FedSV faces a challenge in large scale federated learning in practice -- it may create unfairness.  In order to reduce communication costs, many widely used federated learning algorithms in each round select only a subset of clients to upload their local models~\cite{mcmahan2017communication, nishio2019client}. In the design of FedSV, the unselected clients are assigned with zero credit in that round. This design causes unfairness. One particular example implied by the fundamental requirements of the Shapley fairness~\cite{shapley201617} is that if two clients provide the same data and thus have the identical capability in utility contribution, then they should receive the same rewards. While in the setting of FedSV, clients with the same local data may be assigned with very different credit due to random selection in the training (see an empirical case study in Section~\ref{sec:6.2}). 

The fundamental challenge of fair evaluation of data owners' contributions in federated learning is that we have to retain fairness in pursuing efficiency. In this paper, we develop a principled approach. The general idea is intuitive and effective.  We propose a novel notion of \emph{utility matrix} consisting of contributions of all possible subsets of the clients over all training rounds. Obviously, with the utility matrix, even FedSV can make a fair evaluation on all clients.  However, the utility matrix can only be partially observed because only a subset of clients is selected in each round. The weakness of FedSV is that it computes the Shapley values using only those observed entries in the utility matrix directly. Our idea is try to complete the missing entries of the utility matrix so that the unfairness can thus be eliminated. 

Completing the unobserved entries is far from trivial. We rigorously show that the utility matrix is approximately low-rank based on a careful theoretical analysis of the similarity between contributions from different clients in successive training rounds. This theoretical insight enables us to estimate the missing entries in the utility matrix using techniques from low-rank matrix completion.  With the above insights, we propose a new measure, called \emph{completed FedSV}, to evaluate data owners' contributions in federated learning. 
Our primary technical contributions are as follows.

\begin{enumerate}
    \item We propose the notion of utility matrix and prove that the matrix is approximately low-rank when the local models are Lipschitz continuous and smooth (Propositions~\ref{prop:lipschitz} and~\ref{prop:strongly_cvx}). 
    \item We develop a new measure for fair contribution valuation in federated learning, called completed federated Shapley value (ComFedSV), based on solving a low-rank matrix completion problem for the utility matrix. 
    \item We show that if the utility matrix is well completed, then ComFedSV satisfies two desirable properties for fairness, i.e., symmetry and zero element (Theorem~\ref{prop:fair}).
    \item To tackle the practical scenarios where the size of the utility matrix is too large to complete directly, we adopt a Monte-Carlo sampling method to reduce both space and time complexity (Algorithm~\ref{alg:main}).  
\end{enumerate}

The rest of the paper is organized as follows. Section~\ref{sec:2} reviews the literature on many data valuation techniques for machine learning and federated learning. We introduce the federated learning problem as well as the algorithm for solving it in section~\ref{sec:3}. In section~\ref{sec:fair}, we formalize the definition of fairness for data valuation under the setting of federated learning. Section~\ref{sec:4} introduces the FedSV proposed by Wang \textit{et al.}~\cite{wang2020principled} and shows that it may break the fairness requirement for data valuation. We propose the improved data valuation metric -- ComFedSV in section~\ref{sec:5}. Specifically, section~\ref{sec:5.1} introduces the utility matrix and theoretically shows its low-rank structure, section~\ref{sec:5.2} illustrates the procedure for completing the utility matrix, section~\ref{sec:5.3} formally defines the ComFedSV and section~\ref{sec:5.4} provides an efficient algorithm for estimating the ComFedSV. Section~\ref{sec:6} includes some numerical experiments showing the fairness and effectiveness of the ComFedSV, and the appendix contains the proofs for all the theoretical results. 

\section{Related Work} \label{sec:2}
There are many data valuation strategies in literature, including query and view-based pricing~\cite{koutris2015query, koutris2012querymarket, koutris2013toward} and data quality-based pricing~\cite{heckman2015pricing, pipino2002data}. Limited by space, here we restrict our discussion to the related work on data valuation techniques for machine learning and federated learning, which can be broadly characterized into two categories: Shapley-value-based and non-Shapley-value-based. Please see~\cite{pei2020survey} and \cite{cong2021data} for more thorough surveys on data pricing. 

\subsection{Shapley-value-based Data Valuation}
Shapley value~\cite{shapley201617} has had extensive influence in economics~\cite{gul1989bargaining}. Dubey~\cite{dubey1975uniqueness} showed that Shapley value is the unique measure that satisfies the four fundamental requirements of fairness proposed by Shapley~\cite{shapley201617}. 

Shapley value was introduced into the field of machine learning for feature selection~\cite{cohen2005feature, lundberg2017unified,strumbelj2010efficient}, where Shapley value based measures were developed to evaluate the contribution of different features during the training process and then identify features that are the most influential for the model output. Ghorbani and Zou~\cite{ghorbani2019data} pioneered the use of a Shapely value based measure to quantify data contributions in the machine learning context. They proposed a measure called data Shapley value, which can be used to quantify the contribution of a single data point to a learning task. They pointed out that direct computation of data Shapley value needs exponential time complexity and proposed two heuristic approximation methods to improve efficiency. Jia \textit{et al.}~\cite{jia2019towards} provided several efficient algorithms for approximating the data Shapley value, including group-testing and compressed-sensing based approximation methods. They also investigated the computational complexity for the high probability guarantee on approximation error. Ghorbani \textit{et al.}~\cite{ghorbani2020distributional} extended the notion of data Shapley value from a fixed training data set to arbitrary data distribution, called distributional Shapley value. Theoretically, they proved that their proposed measure is stable in the sense that similar points receive similar values, and similar distributions yield similar value functions. They also provided an algorithm for approximating the distributional Shapley value with a theoretical guarantee on the approximation error. As a follow-up,  Kwon \textit{et al.}~\cite{kwon2021efficient} recently developed analytic expressions for distributional Shapley value for several machine learning tasks, including linear regression and binary classification. They also proposed efficient algorithms for computing the distributional Shapley value based on these analytic expressions. 

Song \textit{et al.}~\cite{song2019profit} extended the notion of data Shapley value to federated learning, called contribution index. They pointed out that directly computing the contribution index needs retraining the model exponentially many times, which is unaffordable in federated learning. As a solution, they proposed two gradient-based heuristics for approximating the contribution index. The key idea is to approximate the local models by leveraging the gradients during the training process of the global model. However, there is no theoretical guarantee on fairness for their proposed methods. 

The need of retraining the model for different subsets of participants is a bottleneck of Shapley value computation in federated learning. Wang \textit{et al.}~\cite{wang2020principled} alternatively solved this problem by proposing a new measure, called federated Shapley value, which can be determined from local model updates in each training iteration, and thus no retraining of the model is required. It also satisfies many desirable properties. However, as discussed in Section~\ref{sec:1}, federated Shapley value may introduce unfairness. We give a more detailed description of federated Shapley value in Section~\ref{sec:4}. 

\subsection{Non-Shapley-value-based Data Valuation}
There are also a few data valuation strategies for machine learning and federated learning that do not depend on Shapley value. Koh and Liang~\cite{koh2017understanding} developed a permutation-based data valuation method for identifying training points most responsible for a given prediction. Their method depends on computing influence functions, which is a classic technique from robust statistics \cite{hampel1974influence}. Wang \textit{et al.}~\cite{wang2019measure} brought the similar idea to horizontal federated learning with a different influence function, where the influence of a client is defined as the sum of the influence of the client's data points. Yoon \textit{et al.}~\cite{yoon2020data} proposed a reinforcement learning-based method to adaptively learn the contribution of each data point towards the learned predictor model. Zhao \textit{et al.}~\cite{zhao2021efficient} recently applied the same technique to federated learning. 

Compared with Shapley-value-based data valuation techniques, the above mentioned non-Shapley-value-based data valuation techniques are usually more computationally efficient as they require less or even no retraining of models. However, non-Shapley-value-based data valuation techniques usually cannot provide any theoretical guarantee on fairness, which is an important desiderata in data valuation~\cite{ghorbani2019data, pei2020survey}. This motivates us to develop efficient data valuation techniques for federated learning that retains fairness.

\section{Federated Learning} \label{sec:3}
In this section, we revisit the federated learning model as well as the algorithm used to optimize the model. Suppose the $i$-th data owner has training data set $D_i$. We consider the following distributed optimization problem.
\begin{equation} \label{eq:main_problem}
    \min_{w}\enspace F(w) := \frac{1}{N}\sum_{i = 1}^N F_i(w),
\end{equation}
where $w\in\mathbb{R}^n$ is the model parameter, $N$ is the number of data owners, and the local objective for client $i$ is
\begin{equation} \label{eq:loss}
    F_i(w) := \ell(w; D_i),
\end{equation}
where $\ell$ is a differentiable loss function. This setting guarantees that if two clients have the identical local data sets, then they have the same local models, i.e., $D_i=D_j$ implies $F_i=F_j$. 

To optimize the distributed optimization problem in Equation~\eqref{eq:main_problem}, we use the federated averaging (\texttt{FedAvg}) method~\cite{mcmahan2017communication}, which is the first and perhaps the most widely used federated learning algorithm. In each round, \texttt{FedAvg} runs several steps of stochastic gradient descent in parallel on a randomly sampled subset of participants and then averages the resulting model updates via a central server. We describe one (say the $t$-th) round of the standard \texttt{FedAvg} algorithm. 

Let $I = \{1, \dots, N\}$ denote the set of all participants. In round $t$, the \texttt{FedAvg} executes the following steps.
\begin{enumerate}
    \item The central server broadcasts the latest model $w^t$ to all data owners;
    \item Every data owner $i$ updates the local model by setting $w_i^t = w^t$ for all $i$;
    \item Every data owner $i$ performs local updates. For all $i$,
    \begin{equation} \label{eq:local_step}
        w_i^{t+1} = w_i^t - \eta^t \nabla F_i(w_i^t),
    \end{equation}
    where $\eta^t$ is the learning rate used in the $t$-th round. 
    \item A subset $I_t \subseteq I$ of participants is randomly selected with uniform probability by the central server;
    \item The central server aggregates the selected local models to produce a new global model
    \begin{equation} \label{eq:global_step}
        w^{t+1} = \frac{1}{|I_t|} \sum_{i \in I_t} w_i^{t+1}.
    \end{equation}
\end{enumerate}

For simplicity, here we let each client do only one step of deterministic local update, that is, $w_k^{t+1} = w_k^t - \eta^t \nabla F_k(w_k^t)$. Our theoretical results can be generalized to an arbitrary number of stochastic local updates in each round. 

\section{Relaxation of Fairness in Data Valuation in Federated Learning} \label{sec:fair}
Fairness is one of the most desirable properties for a data valuation metric \cite{ghorbani2019data,pei2020survey}. In the federated learning setting, a data valuation metric should be able to reflect how much each data owner contributes to the performance of the final model. Formally, we assume a black-box utility function $U:2^I \to \mathbb{R}$ such that for any subset of clients $S \subseteq I$, $U(S)$ returns a utility score of the model collaboratively trained by clients in $S$, such as the performance of the model. Let $v: I \to \mathbb{R}$ be the evaluation metric associated with the utility function $U$. Given the utility function $U$,  the Shapley fairness~\cite{shapley201617} has four fundamental requirements for a metric $v$.
\begin{enumerate}
    \item \textbf{Symmetry.} For any two clients $i, j \in I$, if for any subset of clients $S \subseteq I \setminus \{i,j\}$, $U(S \cup \{i\}) = U(S \cup \{j\})$, then $v(i) = v(j)$. 
    \item \textbf{Zero element.} For any client $i \in I$, if for any subset of clients $S \subseteq I \setminus \{i\}$, $U(S \cup \{i\}) = U(S)$, then $v(i) = 0$.
    \item \textbf{Additivity.} If the utility function $U$ can be expressed as the sum of separate utility functions, namely $U = U_1 + U_2$ for some $U_1, U_2 : 2^I \to \mathbb{R}$, then for any client $i \in I$, $v(i) = v_1(i) + v_2(i)$, where $v_i$ and $v_2$ are the evaluation metrics associated with the utility functions $U_1$ and $U_2$, respectively. 
    \item \textbf{Balance.}  $v(S) = \sum_{i \in S} v(i)$.
\end{enumerate}

Under the federated learning setting, a utility function is usually defined using the model performance on a test data set. The symmetry requires that the same contributions to the utility should receive the same evaluation, which implies that clients with same local data sets should receive same evaluation. The zero element requires that no contribution, no value is recognized. The additivity requires that if there are multiple tasks and thus multiple test data sets, then the contributions of any client with respect to the test data sets can be expressed as the sum of the contributions with respect to those different tasks and test data sets. It is worth noting that although balance is an important requirement in economics, which guarantees that the payment is fully distributed to all players, it is not relevant in the setting of this paper, since we only care about the relative contributions between clients. 

It is shown~\cite{dubey1975uniqueness,ghorbani2019data} that if the data valuation metric $v$ satisfies symmetry, zero element, and additivity, then $v$ must have the form
\begin{equation} \label{eq:shapley}
    v(i) = c \sum\limits_{S \subseteq I \setminus\{i\}} \frac{1}{\binom{N-1}{|S|}} \left[U(S\cup\{i\}) - U(S)\right],
\end{equation}
for some positive constant $c$. However the data valuation metric in form~\eqref{eq:shapley} cannot be directly applied in federated learning, because evaluation of the utility function $U$ requires retraining models \cite{ghorbani2019data}, which is impractical in federated learning \cite{wang2020principled}. Therefore, we cannot expect any practical data valuation metric in federated learning exactly satisfies the above requirements for fairness. 

To tackle this challenge, we propose $\epsilon$-Shapley fairness, a relaxation of fairness for federated learning.

\begin{defi}[$\epsilon$-Shapley fairness] \label{def:fairness}
    Given a utility function $U:2^I \to \mathbb{R}$ and a data valuation metric $v: I \to \mathbb{R}$, $v$ is \emph{$\epsilon$-Shapley-fair} with respect to $U$ for $\epsilon > 0$ if the following two properties hold.
    \begin{enumerate}
    \item \textbf{$\epsilon$-Symmetry.} For any two clients $i, j \in I$, if for any subset of clients $S \subseteq I \setminus \{i,j\}$, $U(S \cup \{i\}) = U(S \cup \{j\})$, then $|v(i) - v(j)| \leq \epsilon$. 
    \item \textbf{$\epsilon$-Zero element.} For any client $i \in I$, if for any subset of clients $S \subseteq I \setminus \{i\}$, $U(S \cup \{i\}) = U(S)$, then $v(i) \leq \epsilon$.
    \item \textbf{$\epsilon$-Additivity.} If the utility function $U$ can be expressed as the sum of separate utility functions, namely $U = U_1 + U_2$ for some $U_1, U_2 : 2^I \to \mathbb{R}$, then for any client $i \in I$, 
    \[|v(i) - (v_1(i)  + v_2(i))| \leq \epsilon,\]
    where $v_1$ and $v_2$ are the evaluation metrics associated with the utility functions $U_1$ and $U_2$, respectively. 
\end{enumerate}
\end{defi}
In $\epsilon$-Shapley fairness, parameter $\epsilon$ is used to control the quantitative requirement on fairness. 

\section{Federated Shapley Value} \label{sec:4}
In this section, we first review the roundly utility function and the FedSV proposed by Wang \textit{et al.}~\cite{wang2020principled}. Then we give a simple example showing that directly using FedSV for data valuation may cause some unfairness. 

Data valuation in federated learning aims to find data sets that are important or influential to the learning task. In federated learning, the importance of a data set is reflected by how much it can improve the performance of the final model. This idea is well established under the general context of machine learning~\cite{ghorbani2019data}. With this insight, we evaluate the importance of the data sets using the test loss of the model. More specifically, consider a typical federated learning process of $T$ rounds as described in Section~\ref{sec:3}. For any $t$ $(1 \leq t \leq T)$, let $u_t: \mathbb{R}^n \to \mathbb{R}$ denote the per-round utility function defined as
\begin{equation} \label{eq:utility}
    u_t(w) = \ell(w^t; D_c) - \ell(w; D_c),
\end{equation}
where $D_c$ is the test data set kept by the central server. Function $u_t(w)$ measures the model performance in round $t$ with model parameter $w$ by the decrease in loss on the test set. 

For round $t $, we define the utility function $U_t : 2^{I} \to \mathbb{R}$. For any subset of participants $S \subseteq I$, the utility created by those participants in round $t$ is defined as
\[U_t(S) := u_t(w_S^{t+1})  \enspace\text{where}\enspace w_S^{t+1} = \frac{1}{|S|}\sum_{k\in S} w_k^{t+1}.\]

The federated Shapley value~\cite{wang2020principled} measures the average marginal utility over the selected clients. 
\begin{defi} \label{def:federated_sv}
    The FedSV $s_{t, i}$ of client $i \in I$ in round $t$ is 
    \begin{equation*}
    s_{t, i} = 
        \begin{cases} 
      \frac{1}{|I_t|} \sum\limits_{S \subseteq I_t \setminus\{i\}} \frac{1}{\binom{|I_t|-1}{|S|}} \left[U_t(S\cup\{i\}) - U_t(S)\right] & i \in I_t \\
      0 & i \notin I_t 
   \end{cases}
    \end{equation*}
    The final FedSV of client $i \in I$ takes the sum of the values of all rounds, that is, 
   $s_i = \sum_{t=1}^T s_{t, i}$.
\end{defi}

If we run only one round of training and select all clients for that round, then FedSV is equivalent to the classical Shapley value.  As mentioned in Section~\ref{sec:1}, more often than not, the central server only selects a subset of clients in each round to reduce communication costs. We argue that simply setting the contribution of an unselected client to zero in a round causes unfairness. One quick example to demonstrate the unfairness is that two clients with the same data may receive quite different evaluations if one of them is unfortunately not selected. We formalize this circumstance in the following observation. 

\begin{obs}[Unfairness of FedSV] \label{obs:unfairness_fedsv}
    Suppose clients $i, j \in I$ have identical local data, that is, $D_i = D_j$, $T$ rounds of the \texttt{FedAvg} algorithm is executed, and the random subsets $I_t \subset I$ of clients satisfying $|I_t| = m < N$. We also assume that the data for clients $i$ and $j$ is useful in the sense that for some $\delta > 0$, for any $t \in \{1, \dots, T\}$, and for any $k \in \{i,j\}$,
    \[s_{t, k} = 
        \begin{cases} 
      \delta_t & k \in I_t \\
      0 & k \notin I_t 
   \end{cases},\]
   where $\delta_t \sim \mathcal{N}(\delta, \sigma_t^2)$ with some $\sigma_t > 0$. It can be easily shown that $\mathbb{E}[s_i] = \mathbb{E}[s_j]$, where $s_i$ and $s_j$, respectively, are the FedSVs for clients $i$ and $j$ as per Definition~\ref{def:federated_sv}. However, the equality in expectation does not guarantee fairness at all, since we only train the model once.
Indeed, $s_i$ and $s_j$ differ from each other with high probability. Specifically, for any $s \in \{0, \dots, T\}$, we have $|s_i - s_j| \geq s\delta$ with probability at least
    \[\mathbb{P}_s := \sum_{a = s}^T \sum_{b=0}^{\floor{\frac{n-a}{2}}} \binom{T}{b,, T - a - 2b, b+a} p^{2b + a} (1-p)^{T - 2b - a}\]
    where $p := \frac{m(N-m)}{N(N-1)}$. This implies that with probability $\mathbb{P}_s$, FedSV is not $s\delta$-Shapley-fair according to Definition~\ref{def:fairness}. The derivation of this probabilistic bound can be found in the appendix. We show a plot of the probability $\mathbb{P}_s$ with different $p$ in Figure~\ref{fig:ps}.
    \begin{figure}
        \centering
        \includegraphics[width=.9\linewidth]{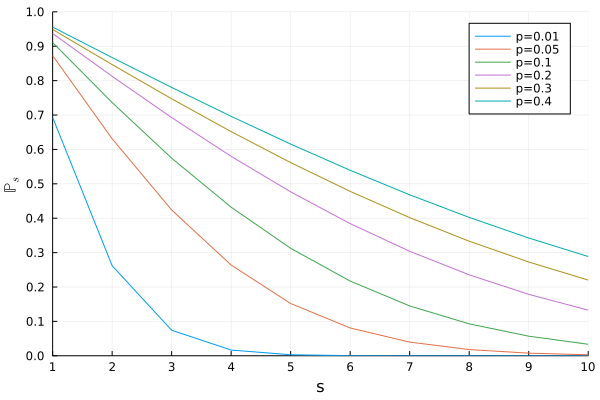}
        \caption{Plot of $\mathbb{P}_s$ described in Observation 1.}
        \label{fig:ps}
    \end{figure}
\end{obs}

The following numerical example corroborates this observation. 
\begin{exam}[Unfairness of FedSV] \label{ex:1}
Consider the MNIST data set~\cite{lecun-mnisthandwrittendigit-2010}. Set the number of clients to be 10 with indices $I = \{0, \dots, 9\}$. Distribute the data for clients $0, \dots, 8$ in a non-IID fashion as described in Section~\ref{sec:6.1} and set client $9$'s data set to be the same as client $0$'s. We train a simple fully connected neural network model using \texttt{FedAvg} for 10 rounds and randomly select 3 clients in each round. We repeat this experiment 50 times. The empirical results show that the relative difference 
\begin{equation} \label{eq:relative_difference}
        d_{0,9} = \frac{|s_0 - s_9|}{\max\{s_0, s_9\}}
\end{equation}
between the FedSVs of $s_0$ and $s_{9}$ is greater than 0.5 with probability 65\%. This simple result clearly shows that, although clients $0$ and $9$ have identical data, they receive dramatically different FEDSVs. FedSV violates the symmetry requirement of fairness.  
\end{exam}

\begin{figure*}[t]
    \centering
    \includegraphics[width=.9\linewidth]{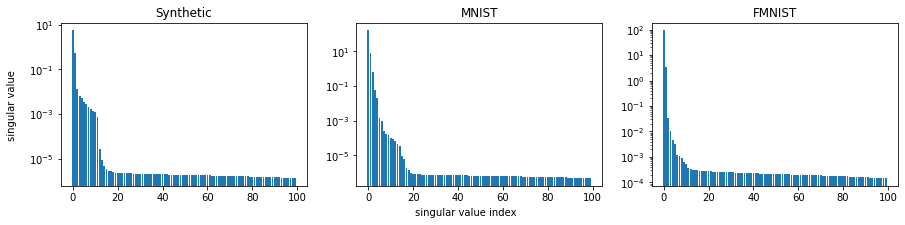}
    \caption{Singular values of the utility matrices described in Example~\ref{ex:2}. In all cases, the matrices are nearly low-rank.}
    \label{fig:svds}
\end{figure*}

\section{Completed Federated Shapley Value} \label{sec:5}
In this section, we first design a utility matrix consisting of contributions of all subsets of clients over all rounds of training, and present the main theoretical result that the utility matrix is approximately low-rank when the local objectives are Lipschitz continuous and smooth. Then, based on a low-rank matrix completion formulation, we propose a new data evaluation method ComFedSV for federated learning. We also show that when the utility matrix can be well recovered, ComFedSV satisfies certain desirable fairness properties. Last, when the size of the utility matrix is too large, we develop a Monte-Carlo type sampling method to reduce the size of the corresponding matrix completion problem.

\subsection{Utility Matrix} \label{sec:5.1}
We design a matrix $\mathcal{U} \in \mathbb{R}^{T \times 2^N}$ that consists of all the possible utilities from any subsets of the clients over all the rounds. Each entry $(t, S)$ is defined as 
$\mathcal{U}_{t, S} = U_t(S)$. If we can compute matrix $\mathcal{U}$, we can obtain the fair Shapley value for every subset of clients in every round.

Since the central server randomly selects a subset $I_t \subseteq I$ of the clients in each round $t$, only a small number of the entries in matrix $\mathcal{U}$ are observed, that is, 
$\{\mathcal{U}_{t, S}: S \subseteq I_t, \enspace t \leq T\}$.
Based on this formulation, a natural question is whether we can get a good approximation for the missing values of the utility matrix $\mathcal{U}$. 

In order to complete a matrix with many missing entries, we have to explore the structure and the properties of the matrix. Our insight into the utility matrix is that, intuitively, it should be approximate low-rank. The rationale comes from at least two reasons. First, the clients with similar data should create similar utilities, which can lead to the similarity between columns of the utility matrix $\mathcal{U}$.  Moreover, if the user-specified loss function $\ell$ does not change dramatically over rounds, then the utilities by the same subset of clients should be similar between successive rounds, which can lead to the similarity between adjacent rows of the utility matrix $\mathcal{U}$.

Before we provide a concrete mathematical justification, let us employ some empirical examples to verify our insight. In the following example, we show that the utility matrix $\mathcal{U}$ is nearly low-rank in many circumstances, in the sense that only a few of its singular values are dominant and all the others are relatively negligible.

\begin{exam}[Low-rankness of Utility Matrix] \label{ex:2}
We use three simple examples to show empirically that the utility matrix is approximately low-rank.  We train a logistic regression model, a fully connected neural network model, and a convolutional neural network model respectively for synthetic data, MNIST~\cite{lecun-mnisthandwrittendigit-2010}, and CIFAR10~\cite{Krizhevsky09learningmultiple} (see Section~\ref{sec:6} for a detailed description). On each data set, we have 10 clients, train the model for 100 rounds and randomly select 3 clients in each round. Thus, the utility matrix $\mathcal{U}$ has size $100\times 2^{10}$. Although we update the global model with the randomly selected clients in each round, in order to obtain the whole utility matrix, we do compute the updates of all clients in each round.

The distribution of singular values is plotted in Figure~\ref{fig:svds}. Clearly, the three different utility matrices are all nearly low-rank.
\end{exam}

Next, we investigate theoretical guarantees to support our intuition. Before stating the result, we first revisit a formal definition of approximately low-rank~\cite{udell2019big}. 
\begin{defi}[$\epsilon$-rank]
    Let $X\in\mathbb{R}^{m\times n}$ be a matrix and $\epsilon>0$ a tolerance. The $\epsilon$-rank of $X$ is given by 
    \[\rank_{\epsilon}(X) = \min\{\rank(Z): Z\in\mathbb{R}^{m\times n}, \|Z - X\|_{\max} \leq \epsilon\},\]
    where $\|\cdot\|_{\max}$ is the absolute maximum matrix entry. Thus, $k = \rank_{\epsilon} (X)$ is the smallest integer for which $X$ can be approximated by a rank $k$ matrix, up to an accuracy of $\epsilon$.
\end{defi}

The following proposition formally characterizes the low-rankness of the utility matrix. We show that if all the losses are Lipschitz continuous and smooth, then the utility matrix is approximately low-rank. The definitions of convex, strongly-convex, Lipschitz, and smooth functions, as well as the proofs of the formal results in this section can be found in the appendices. 

\begin{prop}[Lipschitz continuous and smooth objective] \label{prop:lipschitz}
    Assume that the user specified loss function $\ell(w; D)$ is convex, $L_1$-Lipschitz continuous and $L_2$-smooth in $w$ for all $D \in \{D_c, D_1, D_2, \dots, D_N\}$, and the learning rates $\eta^t$ are non-increasing. 
    Then for any $\epsilon>0$, 
    \begin{equation} \label{eq:epsilon_rank_u}
        \Scale[0.96]{\rank_{\epsilon}(\mathcal{U}) \leq \left\lceil \frac{(2+\eta^1L_2) L_1 \sum_{t=1}^{T-1} \|w^{t} - w^{t+1}\| + (\eta^1 - \eta^T) L_1^2}{\epsilon} \right\rceil.}
    \end{equation} 
\end{prop}

Proposition~\ref{prop:lipschitz} suggests the question: Is there an upper bound on the right-hand side of~\eqref{eq:epsilon_rank_u}? The answer is affirmative, provided an additional assumption on the strong convexity of the loss functions. Li~\textit{et al.}~\cite{li2019convergence} proved that \texttt{FedAvg} converges with rate $\mathcal{O}(1 / T)$ when the local models are all smooth and strongly convex. Thus, we can bound the diameter of the set of global parameters $\{w^1, \dots, w^T\}$. The following result formalizes this observation. 

\begin{prop}[Lipschitz, smooth and strongly convex objective] \label{prop:strongly_cvx}
    In addition to the setting in Proposition~\ref{prop:lipschitz}, we assume that the user specified loss function $\ell(w; D)$ is $\mu$-strongly convex w.r.t.\ $w$ for all $D \in \{D_c, D_1, D_2, \dots, D_N\}$, and the learning rates are defined by 
    \[\eta^t = \frac{2}{\mu(\gamma + t)} \enspace\text{with}\enspace \gamma = \max\left\{\frac{8\mu}{L_2}, 1\right\}.\]
    Then for any $\epsilon>0$, 
    \begin{align*}
        \rank_{\epsilon}(\mathcal{U}) &\leq \left\lceil \frac{2(2+\eta^1L_2) L_1 \log(T)}{\mu\epsilon} +  \frac{(\eta^1 - \eta^T)L_1^2}{\epsilon} \right\rceil
        \\&\in \mathcal{O}\left(\frac{\log(T)}{\epsilon}\right).
    \end{align*}
\end{prop}
This promising result shows that the $\epsilon$-rank of the utility matrix $\mathcal{U}$ is of order $\log(T)$ and is independent of the number of clients. The parameter $\epsilon$ controls the tradeoff between rank and approximation accuracy. 

The sufficient conditions in Propositions~\ref{prop:lipschitz} and~\ref{prop:strongly_cvx} cover a class of Lipschitz continuous, smooth and strongly convex functions, which appear in a wide range of machine learning and optimization problems, including regularized linear regression and logistic regression. This class of functions have many good properties in the sense that neither the function itself nor its gradient has sharp changes. Geometrically, for a set of clients, these two properties can imply the similarity between their contributions over successive rounds, which leads to the low-rank property as stated in Propositions~\ref{prop:lipschitz}and~\ref{prop:strongly_cvx}. For more general loss functions, i.e., neural networks, although the sufficient conditions are not met, we empirically show that the utility matrix is still approximately low rank (Example~\ref{ex:2}).

\subsection{Everyone Being Heard Assumption} 

Propositions~\ref{prop:lipschitz} and~\ref{prop:strongly_cvx}, together with our empirical observation, imply that the utility matrix $\mathcal{U}$ is approximately low-rank when the loss function enjoys certain standard properties. The low-rank structure of the utility matrix suggests that we can approximate the missing entries of $\mathcal{U}$ by solving a low-rank matrix completion problem.

Before describing the low-rank matrix completion problem, we need to introduce a natural assumption. Note that if one client never reports its local model to the server, 
the central server does not get a chance to observe the possible contribution by the client. Consequently, the contribution of the client cannot be estimated. To tackle this issue, we assume that every client is willing to participate in the federated learning in each round.  Therefore, no matter a client is selected by the server in a round, the potential value of the client should be recognized properly.  Formally, we have the following assumption.

\begin{ass}[Everyone Being Heard] \label{ass:main}
The central server selects all clients for update in at least one round, that is, there exists $k>0$, $I_k = I$.  Without loss of generality, in the rest of the paper, we assume $I_1=I$. Our discussion can be easily generalized to any $I_k=I$.
\end{ass}

Note that Assumption~\ref{ass:main} does not come with high costs. As explained in Section~\ref{sec:1}, the central server selects a subset of clients in each round only to reduce communication cost, and the communication cost for doing one round of all clients is equivalent to doing $\left \lceil{N / m}\right \rceil $ rounds of additional training, where $m = |I_t|$ for all $t$. 

\subsection{Approximately Low-Rank Matrix Completion} \label{sec:5.2}

We propose the following factorization-based low-rank matrix completion problem to complete the utility matrix $\mathcal{U}$: 
\begin{equation}
\label{prob:matrix_completion}
\Scale[0.83]{\minimize{\substack{W \in \mathbb{R}^{T \times r}\\ H \in \mathbb{R}^{2^N \times r}}} \enspace \sum_{t=1}^T\sum_{S\subseteq I_t} (\mathcal{U}_{t,S} - w_t^Th_{S})^2 + 
\lambda(\|W\|_F^2 + \|H\|_F^2),}
\end{equation}
where $r$ is a user specified rank parameter, $\lambda$ is a positive regularization parameter, $\|\cdot\|_F$ is the Frobenious norm, and $w_t$ and $h_S$, respectively, are the $t$-th and the $S$-th row vectors of the matrices $W$ and $H$. Because the matrix completion problem~\eqref{prob:matrix_completion} is solved after the whole training process, we can determine the rank parameter $r$ via Propositions~\ref{prop:lipschitz} and~\ref{prop:strongly_cvx}. We show the impact of rank parameter $r$ on the performance of the low-rank matrix completion problem~\eqref{prob:matrix_completion} via a numerical example.

\begin{exam}[Impact of Rank Parameter]
We conduct a numerical experiment to show the impact of the rank parameter $r$ on the matrix completion problem~\eqref{prob:matrix_completion}. We train a simple multilayer perceptron neural network for the MNIST data set. We have 10 clients, train the model for 100 rounds and randomly select 3 clients in each round. Note that although we update the global model with the selected clients, in order to get the whole utility matrix, we compute the updates of all clients in each round. We define the relative difference between $\mathcal{U}$ and $WH^T$ as 
$\frac{\|\mathcal{U} - WH^T\|_{Fro}}{\|\mathcal{U}\|_{Fro}}$.
We try different rank parameters $r \in \{1, \dots, 10\}$ and report the relative difference in Figure~\ref{fig:rank}. The curve shows that a small rank parameter may not be enough for recovering the information in the utility matrix $\mathcal{U}$ and a large rank parameter may bring some overfitting problem.  

\begin{figure}
    \centering
    \includegraphics[width=.9\linewidth]{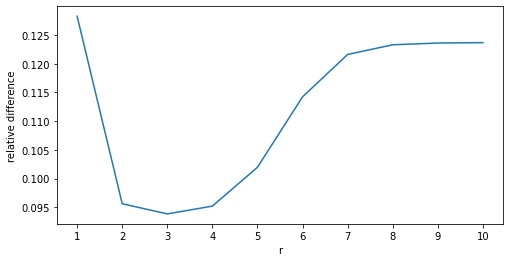}
    \caption{Impact of rank parameter described in Example 3.}
    \label{fig:rank}
\end{figure}

\end{exam}

Interestingly, the low-rank matrix completion model~\eqref{prob:matrix_completion} was first used in completing the information for recommender systems~\cite{koren2009matrix}. Its effectiveness has been extensively studied both theoretically and empirically~\cite{keshavan2012efficient, sun2016guaranteed}.  Therefore, we can simply adopt the well established matrix completion methods~\cite{yu2014parallel, chin2016libmf} and focus on how to derive fair contribution evaluation metric based on a completed utility matrix.

\begin{figure*}
    \centering
    \includegraphics[width=\linewidth]{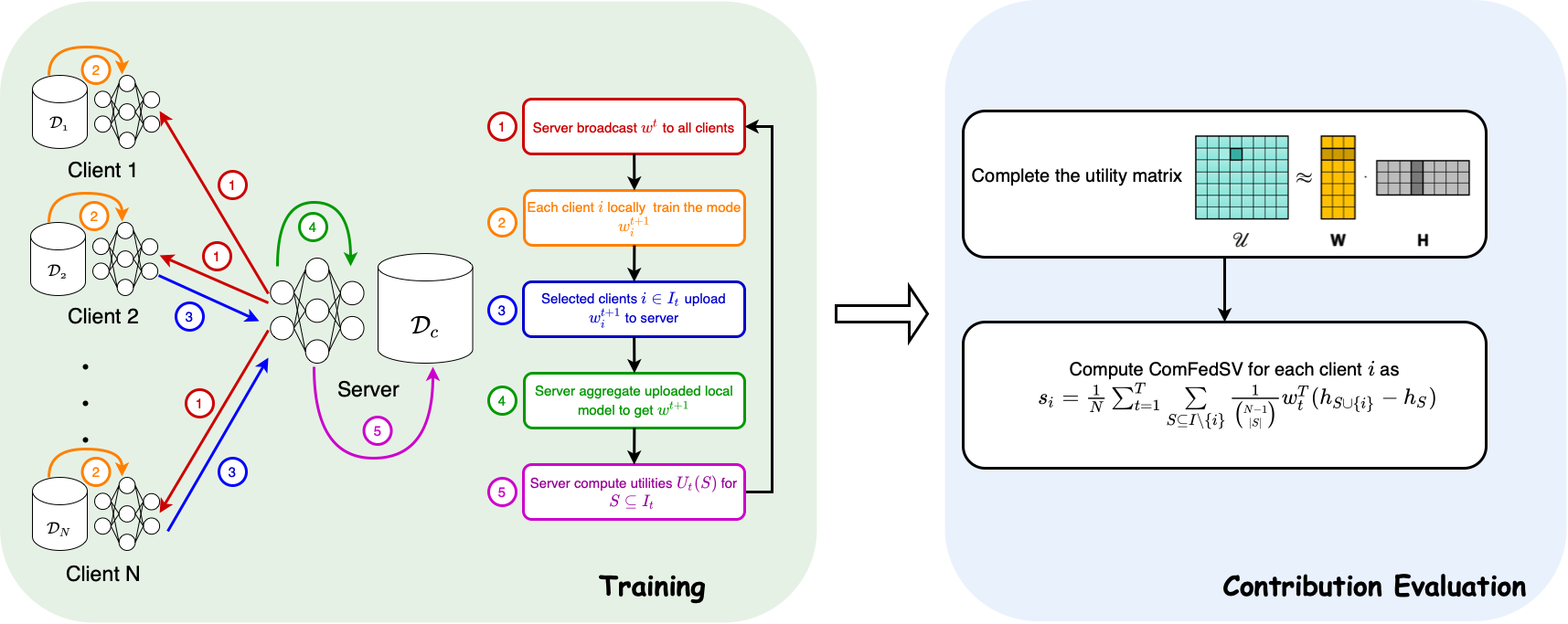}
    \caption{Pipeline of data valuation in horizontal federated learning.}
    \label{fig:pipeline}
\end{figure*}

\subsection{Completed Federated Shapley Value} \label{sec:5.3}
Now, we are ready to give the formal definition of ComFedSV. 
\begin{defi}[Completed Federated Shapley Value (ComFedSV)] \label{def:complete_sv}
    Let $W$ and  $H$ be the optimal solutions to the approximately low-rank matrix completion problem~\eqref{prob:matrix_completion}. Then, the ComFedSV for each client $i \in I$ is defined as 
    \begin{equation} \label{eq:complete_sv}
        s_i = \frac{1}{N} \sum_{t=1}^T\sum\limits_{S \subseteq I \setminus\{i\}} \frac{1}{\binom{N-1}{|S|}} w_t^T(h_{S\cup\{i\}} - h_S),
    \end{equation}
    where $w_t$ and $h_S$, respectively, are the $t$-th and the $S$-th row vectors of the matrices $W$ and $H$.
\end{defi}
The whole pipeline is shown in Figure~\ref{fig:pipeline}. The next result shows that if we can approximately recover the utility matrix, then ComFedSV is guaranteed to be approximate Shapley-fair according to Definition~\ref{def:fairness}. We first give a formal definition of the approximate recovery of the utility matrix. 
\begin{defi} \label{def:completeness}
     Let $W$ and  $H$ be the optimal solutions to the approximately low-rank matrix completion problem~\eqref{prob:matrix_completion}, and $s$ be the ComFedSV (Definition~\ref{def:complete_sv}). We say that $s$ is \textbf{$\delta$-completed} for some $\delta>0$, if we can complete the utility matrix with tolerance $\delta$, that is,
     $\|\mathcal{U} - W H^T\|_1 \leq \delta$, where
     $\|\cdot\|_1$ is the maximum absolute column sum, i.e., $\|X\|_1 = \max_j \sum_i |X_{i,j}|$. 
\end{defi}
We now state our theoretical guarantee on the fairness of the ComFedSV. 
\begin{theorem}[Fairness Guarantee] \label{prop:fair}
    Let $U: 2^I \to \mathbb{R}$ be the utility function for the whole training process in federated learning: 
    \[U(S) = \sum_{t=1}^T U_t(S) \enspace \forall S \subseteq I.\]
    Let $s$ be the ComFedSV computed with respect to $U$.  Then, $s$ is $(\frac{4\delta}{N})$-Shapley-fair if
 \begin{enumerate}
\item  $s$ is $\delta$-completed for some $\delta > 0$; and 
\item if $U = U_1 + U_2$ and $s_1, s_2$ are the ComFedSVs computed with respect to utility functions $U_1$ and $U_2$, respectively, then $s_i$ $(i \in \{1,2\})$ is $\delta_i$-completed such that $\delta_1 + \delta_2 \leq \delta$. 
\end{enumerate}
\end{theorem}
In particular, as long as the utility matrix is perfectly recovered, i.e., $\mathcal{U} = WH^T$, Theorem~\ref{prop:fair} guarantees that clients with identical local data obtain the same evaluation. 

There are different kinds of fairness in machine learning and federated learning~\cite{zemel2013learning,donini2018empirical,li2019fair,lyu2020collaborative,chu2021fedfair}. In this work, we focus on the fairness in data valuation (Theorem~\ref{prop:fair}). Moreover, our proposed metric is adaptable to different training algorithms for federated learning. When there are training algorithms that can achieve some other requirements on fairness besides the ones described in the paper, it is possible to combine those training algorithms with our valuation metric. For example, Chu et al.~\cite{chu2021fedfair} proposed an algorithm that can guarantee the fairness for protected groups. Our valuation metric ComFedSV can be easily adapted to their algorithm since we still have access to the uploaded local models in each iteration.

\subsection{Estimating ComFedSV} \label{sec:5.4}

\RestyleAlgo{ruled}
\SetKwInput{KwInput}{Input}
\SetKwInput{KwReturn}{Return}
\begin{algorithm}[t]
\DontPrintSemicolon
\caption{Estimating the ComFedSV} \label{alg:main}
\KwInput{$I$: set of clients\\
    \hspace{9mm} $T$: number of rounds\\
    \hspace{9mm} $K$: number of selected clients per round\\
    \hspace{9mm} $M$: number of Monte Carlo samples
    }
Initialize utility matrix $\mathcal{U} \in \mathbb{R}^{T\times MN}$ with all $0$'s\;
Initialize the global model parameter $w^0 \in \mathbb{R}^n$\;
Uniformly sample $M$ permutations of $I$: $\pi_1, \dots, \pi_M$\;
 \For{t = 0, \dots, T}{
    
    \For{$i$ in $I$}{
        Synchronize the global model $w_i^t = w^t$\;
        Update the local model as in~\eqref{eq:local_step}\;
    }
    
    \eIf{t = 0}{
        $I_t = I$; see Assumption~\ref{ass:main}\;
    }{
        Uniformly sample $I_t \subset I$ with $|I_t| = K$\;
    }
    
    Update the global model as in~\eqref{eq:global_step}\;
    
    \For{m = 1, \dots, M}{
        \For{$i$ in $I$}{
            \If{$\pi_m(i) \subseteq I_t$}{
                $\mathcal{U}_{t,\pi_m(i)} = u_t(\pi_m(i))$;
            }
        }
    }
 }
 Solve problem~\eqref{prob:matrix_completion_reduced} to get $W$ and $H$\;
 \For{$i$ in $I$}{
    Compute the completed Shapley value $s_i$ as in~\eqref{eq:complete_ev_approx}\;
 }
 \KwReturn{ $\{s_i: i \in I\}$ }
\end{algorithm}

The utility matrix $\mathcal{U}$ has size $T \times 2^N$, which is exponential in the number of clients. As the number of clients increases, solving the matrix completion problem~\eqref{prob:matrix_completion} cannot scale up. The similar situation appears in the computation of the Shapley value, where the computational complexity is also exponential in the number of clients. Can we estimate the Shapley values in our model?

How to efficiently estimatie the Shapley value has been studied extensively~\cite{ghorbani2019data,jia2019towards}. Here we describe the well known Monte-Carlo sampling method~\cite{metropolis1949monte, ghorbani2019data}. We can rewrite the definition of ComFedSV~\eqref{eq:complete_sv} into an equivalent formulation using expectation:
\begin{equation} \label{eq:complete_ev_expectation}
    s_i = \mathop{\mathbb{E}}_{\pi \sim \Pi(I)} \sum_{t=1}^T w_t^T\left(h_{ \pi(i) \cup \{i\}}  - h_{ \pi(i) }\right),
\end{equation}
where $\Pi(I)$ is the uniform distribution over all $N!$ permutations of the set of clients $I$ and $\pi(i)$ is the set of clients preceding client $i$ in permutation $\pi$. With this formulation, we can use the Monte-Carlo method to get an approximation of ComFedSV. More precisely, we can randomly sample $M$ permutations $\pi_1, \dots, \pi_M$ and get an approximation to ComFedSV $s_i$ by 
\begin{equation} \label{eq:complete_ev_approx}
    \hat s_i = \frac{1}{M}\sum_{m=1}^M \sum_{t=1}^T w_t^T\left(h_{ \pi_m(i) \cup \{i\}}  - h_{ \pi_m(i) }\right).
\end{equation}

From~\eqref{eq:complete_ev_approx}, we notice that we only need access to part of the utility matrix $\mathcal{U}$, namely 
\[ \{\mathcal{U}_{t, S}: t \in [1, \dots, T], S = \pi_m(i), m \in [1,\dots,M], i \in I\}, \]
which suggests that we can reduce the size of the matrix completion problem~\eqref{prob:matrix_completion}. Formally, we can solve the following reduced matrix completion problem:
\begin{align}
\label{prob:matrix_completion_reduced}
\minimize{\substack{W \in \mathbb{R}^{T \times r}\\ H \in \mathbb{R}^{MN \times r}}} &\enspace \sum_{t=1}^T\sum_{ \substack{m \in [1,\dots, M] \\ i \in I \\ \pi_m(i) \subseteq I_t  } } (\mathcal{U}_{t,\pi_m(i)} - w_t^Th_{\pi_m(i)})^2 \nonumber\\
&+ \lambda(\|W\|_F^2 + \|H\|_F^2).
\end{align}

We summarize the whole process in Algorithm~\ref{alg:main}. It was shown by \cite{maleki2013bounding} that, for bounded utility functions, the Monte-Carlo method requires $M = \mathcal{O}(N\log N)$ samples to achieve a good approximation.  The estimation is unbiased. As a consequence, both the space and time complexities for Algorithm~\ref{alg:main} are $\mathcal{O}(TN^2\log(N))$.

\section{Experiments} \label{sec:6}
This section reports a series of experiments that examine how well ComFedSV can capture contributions from clients/data owners in federated learning. Our code is publicly available at the \emph{Huawei AI Gallery}\footnote{\url{https://developer.huaweicloud.com/develop/aigallery/notebook/detail?id=08da32bb-9b85-4ddb-ae6c-94a3f5102133}}.

\begin{figure}[t]
    \centering
    \includegraphics[width=.9\linewidth]{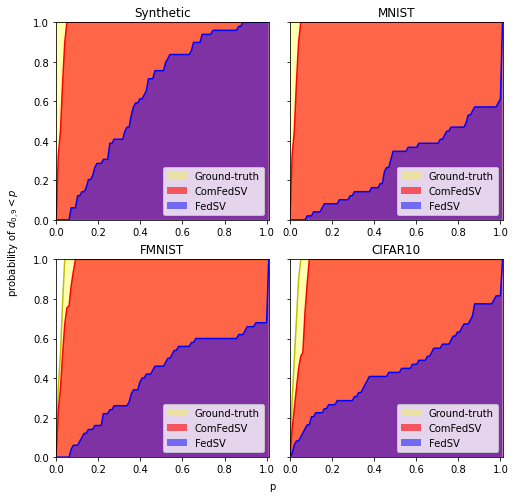}
    \caption{Fairness: the empirical cumulative distribution of the relative difference between the evaluation metrics (FedSV and ComFedSV) of clients 0 and 9, which are set to have the identical local data sets.}
    \label{fig:fireness}
\end{figure}

\begin{figure}[t]
    \centering
    \includegraphics[width=.9\linewidth]{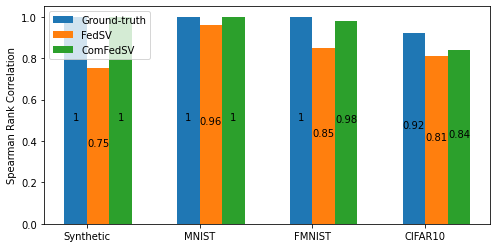}
    \caption{Noisy data detection: the Spearman’s  rank  correlation between the true ranking and the rankings produced by the three metrics (ground-truth, FedSV and ComFedSV).}
    \label{fig:noisy_data_detection}
\end{figure}

\begin{figure*}
     \centering
     \includegraphics[width=.85\linewidth]{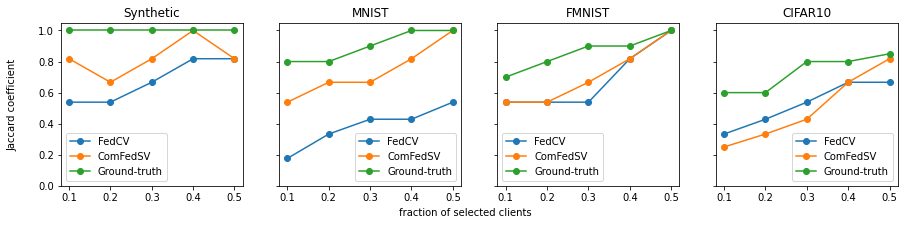}
     \caption{Noisy label detection: the Jaccard coefficient between the set of 10 clients with noisy labels and the set of 10 clients with the lowest evaluations (FedSV and ComFedSV).}
     \label{fig:noisy_label_detection}
\end{figure*}

\subsection{Experiment Setting} \label{sec:6.1}
\subsubsection{Data Sets} We perform experiments on both real and synthetic data. For synthetic data, we follow the setup as in~\cite{li2018federated}. For real data, we choose the widely used benchmark data sets MNIST~\cite{lecun-mnisthandwrittendigit-2010}, Fashion MNIST~\cite{xiao2017fashion}, and CIFAR10~\cite{krizhevsky2009learning}.  

In reality, data sets from different data owners may have different distributions. To model this circumstance, for each data set, we consider two different ways for distributing the data: IID and non-IID. For synthetic data, according to \cite{li2018federated}, the data set has two parameters $\alpha$ and $\beta$, where $\alpha$ controls how much local models differ from each other and $\beta$ controls how much local data sets differ from each other. We choose $\alpha=0, \beta=0$ for IID and $\alpha=1, \beta=1$ for non-IID. For real data sets MNIST, Fashion MNIST, and CIFAR10, we randomly distributed the data to all clients for IID, and each client gets samples of only two classes for non-IID, which is the same setting as in the original \texttt{FedAvg} paper~\cite{mcmahan2017communication}.

\subsubsection{Model} For synthetic data, we train a logistic regression model. We run 100 rounds of training and can achieve up to $99\%$ and $95\%$ classification accuracies on the test set in the cases of IID and non-IID, respectively. For MNIST, we train a fully connected neural network. We run 100 rounds of training and can achieve up to $98\%$ and $92\%$ classification accuracies on the test set in the IID and non-IID settings, respectively. For Fashion MNIST, we train a simple convolutional neural network. We run 100 rounds of training and can achieve up to $98\%$ and $90\%$ classification accuracies on the test set in the cases of IID and non-IID, respectively. For CIFAR10, we train a VGG16 model~\cite{simonyan2014very}, which is a more complex convolutional neural network. We run 100 rounds of training and can achieve up to $91\%$ and $83\%$ classification accuracies on the test set in the cases of IID and non-IID, respectively, where we use a pre-trained model for non-IID.  We use different models to test the generality of our method. 

\subsubsection{Solver for Matrix Completion} We use the open source package \texttt{LIBPMF}\footnote{\url{https://github.com/cjlin1/libmf}}~\cite{chin2016libmf} to solve the matrix completion problem~\eqref{prob:matrix_completion_reduced}.


\subsubsection{Compared metrics} We compare the performance of ComFedSV with FedSV and the ground-truth, where the ground-truth is ComFedSV computed from the full utility matrix. Note that for the ground-truth, although we compute the updates of all clients in each round to obtain the whole utility matrix, we only update the global model with the updates by the randomly selected clients in each round, so that the global models will be the same for all three metrics. 

\subsection{Fairness} \label{sec:6.2}
In this experiment, we extend the results shown in Example~\ref{ex:1} and investigate whether ComFedSV improves fairness by checking if clients with similar data sets receive similar ComFedSVs. Limited by space, we only discuss the non-IID setting here.

\begin{figure*}
     \centering
     \includegraphics[width=.9\linewidth]{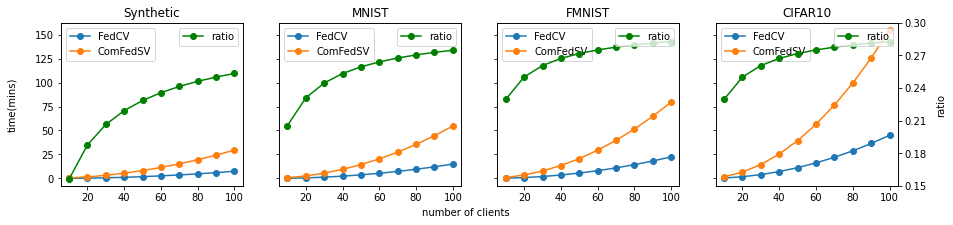}
     \caption{Computing time comparison. The orange and blue curves correspond to the computing time for FedSV and ComFedSV respectively (left y-axis). The green curve the the ratio between the computing time for FedSV and ComFedSV (right y-axis).}
     \label{fig:time_performance}
\end{figure*}

Figure~\ref{fig:fireness} shows the empirical cumulative distribution of the relative difference $d_{0,9}$ between clients 0 and 9, where we use two metrics: FedSV and ComFedSV. The relative difference is defined by Equation~\eqref{eq:relative_difference}, and takes values between $0$ and $1$. The vertical axis represents the probability $\mathbb{P}(d_{0,9} \leq t)$. Clearly, ComFedSV outperforms FedSV in fairness. On all data sets, Synthetic, MNIST, FMNIST, and CIFAR10, the probability $\mathbb{P}(d_{0,9}^{ComFedSV} < t)$ is always greater than or equal to $\mathbb{P}(d_{0,9}^{FedSV} < t)$ for all possible value of $t$. In other words, in most trials, clients with same local data receive more similar evaluations in ComFedSV, but receive quite different evaluations in FedSV. 

\subsection{Data Quality Detection} \label{sec:6.3}

Wang~\textit{et al.}~\cite{wang2020principled} showed that FedSV can be used to detect data quality.  To make a fair comparison, we conduct a similar experiment to compare the data quality detection capability of ComFedSV and FedSV.

Typically, two types of noise, data noise and label noise, is considered. The literature usually treats the percentage of noisy data points and noisy labels as the measures for the quality of data sets~\cite{liebchen2008data, karimi2020deep}. 
We start from the IID partitioning of the data to clients and then add different amounts of noise to different clients. After the noise is added, the data used in federated learning is not IID anymore.

\subsubsection{Noisy Data Detection}
We set 10 clients: $I = \{0, 1, \dots, 9\}$, run federated learning for 10 rounds and randomly select 3 clients to report in each round. For each client $i \in I$, we artificially add Gaussian noise to $5\times i\%$ of their data. Under this setting, the descending order of clients in the amount of noise added is $9, 8, \dots, 0$.  We compute the ground-truth, FedSV, and ComFedSV for all clients and compare the rankings produced by those three metrics. We use Spearman's rank correlation~\cite{zar1972significance,zar2005spearman} to measure the similarity between the true ranking $9, 8, \dots, 0$ and the rankings produced by the three metrics. The result is shown in Figure~\ref{fig:noisy_data_detection}. ComFedSV outperforms FedSV and almost matches the ground-truth on all four data sets. This experiment clearly shows that ComFedSV has a strong capability in detect data quality for machine learning models, even better than that of FedSV.

\subsubsection{Noisy Label Detection}
We set 100 clients such that 10 of them each has 30\% noisy labels, where noise is introduced by random flipping. For each data set, we conduct federated learning for 100 rounds and randomly select $m\%$ clients to report in each round with $m \in \{10, 20 , \dots, 50\}$. We compare FedSV and ComFedSV by the Jaccard coefficient~\cite{jaccard1912distribution} between the set of  clients with the noisy labels and the set of 10 clients with the lowest evaluations. The result is shown in Figure~\ref{fig:noisy_label_detection}. ComFedSV outperforms FedSV in most cases. Both methods perform better as more clients are selected to report in each round.

\subsection{Time Complexity} \label{sec:time}
In this experiment, we compare the time required for computing ComFedSV and FedSV. Although the time complexity for Algorithm~\ref{alg:main} is $\mathcal{O}(TN^2\log(N))$, the main cost is evaluating the roundly utility function $u_t$, which requires computing the test loss of the model. It is easy to see that in Algorithm~\ref{alg:main}, the number of calls to $u_t$ is in $\mathcal{O}(TNKlog(N))$, where $K$ is the number of selected clients per round. As a comparison, computing FedSV with Monte-Carlo requires $\mathcal{O}(TK^2log(K))$ calls. This suggests that the ratio between the number of calls for FedSV and ComFedSV is $\mathcal{O}(\frac{K}{N})$, which equals to the random participation rate.

We set the number of clients to $10, 20, \dots, 100$ and randomly select $30\%$ of clients to report in each round. The results are shown in Figure~\ref{fig:time_performance}, where we plot the computing time for FedSV and ComFedSV (left y-axis), as well as their ratio (right y-axis). As we can see from the plots, the ratio between the computing time for FedSV and ComFedSV is approaching a stable constant close to the participation rate as the number clients increases, which agrees with our complexity analysis.

\section{Conclusion and Future Directions}
In this paper, we tackle the challenging problem of data valuation in horizontal federated learning. We point out the possible unfairness in the existing data valuation method. Then, we propose a new measure ComFedSV, which is not only fair with theoretical guarantee, and also can be evaluated efficiently. We verify the fairness of our measure both theoretically and experimentally. 

Our study advances the frontier in this emerging important direction. There are also many interesting future directions.  For example, it is interesting to extend our methodology to vertical federated learning. As another example, under the framework of federated learning, there may exist some other properties needed for fairness in federated learning in addition to those in the Shapley fairness. 

\appendices
\section{Derivation of Observation 1}
In this section, we show the derivation of the probabilistic bound in Observation~\ref{obs:unfairness_fedsv}. For any $t \in \{1, \dots, T\}$, we define $\alpha_t := s_{t,i} - s_{t,j}$. Then $|s_i - s_j| = |\sum_{t=1}^T \alpha_t|$. By the assumption in Observation~\ref{obs:unfairness_fedsv} and definition of FedSV (Definition~\ref{def:federated_sv}), it follows that $\alpha_t$ can be expressed as 
\[\alpha_t = 
\begin{cases} 
      \delta_t & i \in I_t ~\text{and}~ j \notin I_t \\
      0 & i,j \in I_t ~\text{or}~ i,j \notin I_t\\
      -\delta_t & i \notin I_t ~\text{and}~ j \in I_t. 
\end{cases}
\]
We define some auxiliary variables $\beta_t$ for $t \in \{1, \dots, T\}$ as 
\[\beta_t = 
\begin{cases} 
      \delta & i \in I_t ~\text{and}~ j \notin I_t \\
      0 & i,j \in I_t ~\text{or}~ i,j \notin I_t\\
      -\delta & i \notin I_t ~\text{and}~ j \in I_t. 
\end{cases}
\]
Then we derive the probabilistic bound for $\sum_{t=1}^T \beta_t$. For any $s \in \{1, \dots, T\}$,
\begin{small}
\begin{align*}
    &\mathbb{P}\left(\sum_{t=1}^T \beta_t \geq s\delta\right) 
    \\=&\sum_{a = s}^T \mathbb{P}\left(\sum_{t=1}^T \beta_t = a\delta\right)
    \\=&\sum_{a = s}^T \sum_{b=0}^{\floor{\frac{n-a}{2}}} \binom{T}{b,, T - a - 2b, b+a} p^{2b + a} (1-p)^{T - 2b - a},
\end{align*}
\end{small}
where
\[p = \mathbb{P}(i \in I_t ~\text{and}~ j \notin I_t) = \frac{m(n-m)}{n(n-1)}.\]
Finally, we derive the probability bound of $|s_i - s_j| \geq s\delta$ for any $s \in \{1, \dots, T\}$:
\begin{small}
\begin{align*}
    &\mathbb{P}(|s_i - s_j| \geq s\delta) 
    \\=&\mathbb{P}\left(|\sum_{t=1}^T \alpha_t| \geq s\delta\right)
    \\=&2\mathbb{P}\left(\sum_{t=1}^T \alpha_t \geq s\delta\right)
    \\\geq&2\mathbb{P}\left(\sum_{t=1}^T \alpha_t \geq s\delta ~\bigg\vert~ \sum_{t=1}^T \beta_t \geq s\delta \right) \mathbb{P}\left(\sum_{t=1}^T \beta_t \geq s\delta\right)
    \\=& \sum_{a = s}^T \sum_{b=0}^{\floor{\frac{n-a}{2}}} \binom{T}{b,, T - a - 2b, b+a} p^{2b + a} (1-p)^{T - 2b - a},
\end{align*}
\end{small}
where the last equality follows from the fact 
\[\mathbb{P}\left(\sum_{t=1}^T \alpha_t \geq s\delta ~\bigg\vert~ \sum_{t=1}^T \beta_t \geq s\delta \right) = \frac{1}{2}.\]

\section{Definitions of Several Classes of Functions}
In this section, we introduce several special classes of functions that are considered in our paper. 

\begin{defi}
Consider a differentiable function $f: \mathbb{R}^n \to \mathbb{R}$. Then we have the following definitions:
\begin{itemize}
    \item $f$ is \textbf{convex} if for any $x, y \in \mathbb{R}^n$, $f$ satisfies
    \[f(y) \geq f(x) + \langle \nabla f(x), y-x \rangle;\]
    \item $f$ is \textbf{$\mu$-strongly convex} for some $\mu > 0$ if for any $x, y \in \mathbb{R}^n$, $f$ satisfies
    \[f(y) \geq f(x) + \langle \nabla f(x), y-x \rangle + \frac{\mu}{2}\|x - y\|_2^2;\]
    \item $f$ is \textbf{$L_1$-Lipschitz} for some $L_1>0$ if for any $x, y \in \mathbb{R}^n$, $f$ satisfies
    \[|f(x) - f(y)| \leq L_1\|x-y\|_2;\]
    \item $f$ is \textbf{$L_2$-smooth} for some $L_2>0$ if for any $x, y \in \mathbb{R}^n$, $f$ satisfies
    \[\|\nabla f(x) - \nabla f(y)\|_2 \leq L_2\|x-y\|_2.\]
\end{itemize}
\end{defi}

\section{Proof for Proposition 1}
In this section, we provide the proof for Proposition 1. 
\begin{proof}
    First, we bound the difference between utilities by the same subset of clients over successive rounds. Specifically, for any $t\in\{1,\dots,T-1\}$ and $S \subseteq I$, we have 
    \begin{small}
    \begin{align*}
        &|U_{t}(S) - U_{t+1}(S)|\\
        = &|\ell(w^{t}; D_c) - \ell(w_S^{t+1}; D_c) - \ell(w^{t+1}; D_c) + \ell(w_S^{t+2}; D_c)|
        \\\leq &|\ell(w^{t}; D_c) - \ell(w^{t+1}; D_c)| + |\ell(w_S^{t+1}; D_c) - \ell(w_S^{t+2}; D_c)|
        \\\leq &L_1\|w^{t} - w^{t+1}\| + L_1\|w_S^{t+1} - w_S^{t+2}\|
        \\= &L_1\|w^{t} - w^{t+1}\| + L_1\bigg\|w^{t} - \frac{\eta^t}{|S|}\sum_{i\in S} \nabla F_i(w^{t})
        \\&  - w^{t+1} + \frac{\eta^{t+1}}{|S|}\sum_{i\in S} \nabla F_i(w^{t+1}) \bigg\|
        \\\leq &2L_1\|w^{t} - w^{t+1}\| + \frac{L_1\eta^{t+1}}{|S|}\sum_{i\in S}\left\|\nabla F_i(w^{t}) - \nabla F_i(w^{t+1})\right\| 
        \\&+ \frac{L_1(\eta^{t} - \eta^{t+1})}{|S|}\sum_{i\in S}\left\|\nabla F_i(w^{t})\right\|
        \\\leq &(2+\eta^1L_2) L_1\|w^{t} - w^{t+1}\| + (\eta^{t} - \eta^{t+1})L_1^2.
    \end{align*}
    \end{small}
    Then we can get an upper bound on the sum of all maximum entry-wise norm between successive rows of the utility matrix, namely
    \begin{small}
    \begin{align*}
        &~\sum_{t=1}^{T-1} \left\|\mathcal{U}[t,:] - \mathcal{U}[t+1,:]\right\|_{\max} \\
        \leq &~\sum_{t=1}^{T-1}\left[ (2+\eta^1L_2) L_1\|w^{t} - w^{t+1}\| + (\eta^{t} - \eta^{t+1})L_1^2 \right]
        \\= &~(2+\eta^1L_2) L_1 \sum_{t=1}^{T-1} \|w^{t} - w^{t+1}\| + (\eta^1 - \eta^T)L_1^2.
    \end{align*}
    \end{small}
    Therefore, it follows that 
    \begin{equation*}
        \Scale[0.97]{\rank_{\epsilon}(\mathcal{U}) \leq \left\lceil \frac{(2+\eta^1L_2) L_1 \sum_{t=1}^{T-1} \|w^{t} - w^{t+1}\| + (\eta^1 - \eta^T)L_1^2}{\epsilon} \right\rceil.}
    \end{equation*}
\end{proof}

\section{Proof for Proposition 2}
In this section, we provide the proof for Proposition 2. 
\begin{proof}
 By \cite[Theorem~1]{li2019convergence}, we know that the based on the setting in Proposition 2, the \texttt{FedAvg} can have sublinear convergence rate. Now we will give an upper bound on the length of path of the global parameters $w^t$: 
 \[\sum_{t=1}^{T-1} \|w^{t+1} - w^t\| = \sum_{t=1}^{T-1} \eta^t \leq \sum_{t=1}^{T-1} \frac{2}{\mu t} \leq \frac{2}{\mu} \log(T).\]
 Now combine with the result in Proposition 1, we have the following upper bound on the $\epsilon$-rank of the utility matrix $\mathcal{U}$:
 \[\rank_{\epsilon}(\mathcal{U}) \leq \left\lceil \frac{2(2+\eta^1L_2) L_1 \log(T)}{\mu\epsilon} +  \frac{(\eta^1 - \eta^T)L_1^2}{\epsilon} \right\rceil.\]
\end{proof}

\section{Proof for Theorem 1}
In this section, we provide the proof for Theorem 1. 
\begin{proof}
We denote the ComFedSV computed from the true utility matrix $\mathcal{U}$ as 
\begin{equation}
    \hat s_i = \frac{1}{N} \sum_{t=1}^T\sum\limits_{S \subseteq I \setminus\{i\}} \frac{1}{\binom{N-1}{|S|}} \left[ U_t(S \cup\{i\}) - U_t(S)\right].
\end{equation}

First, we show that $\hat s_i$'s strictly satisfy the \textbf{Symmetry}, \textbf{Zero element} and \textbf{Additivity} properties.
\begin{itemize}
    \item \textbf{Symmetry.} For any two clients $i,j \in I$. Suppose that for any subset of clients $S \subseteq I \setminus \{i,j\}$, $U(S \cup \{i\}) = U(S \cup \{j\})$. Then, 
    \begin{small}
    \begin{align*}
        \hat s_i = &~ \frac{1}{N} \sum_{t=1}^T\sum\limits_{S \subseteq I \setminus\{i\}} \frac{1}{\binom{N-1}{|S|}} \left[ U_t(S \cup\{i\}) - U_t(S)\right] \\
        = &~\frac{1}{N}\sum\limits_{S \subseteq I \setminus\{i\}} \frac{1}{\binom{N-1}{|S|}} \left[\sum_{t=1}^T U_t(S \cup\{i\}) - \sum_{t=1}^T U_t(S)\right]\\
        = &~\frac{1}{N}\sum\limits_{S \subseteq I \setminus\{i,j\}} \biggl\{\frac{1}{\binom{N-1}{|S|}} \left[\sum_{t=1}^T U_t(S \cup\{i\}) - \sum_{t=1}^T U_t(S)\right]  
        \\& + \frac{1}{\binom{N-1}{|S|+1}} \left[\sum_{t=1}^T U_t(S \cup\{j\}\cup\{i\}) - \sum_{t=1}^T U_t(S\cup\{j\})\right]\biggr\}\\
        = &~ \frac{1}{N}\sum\limits_{S \subseteq I \setminus\{i,j\}} \biggl\{\frac{1}{\binom{N-1}{|S|}} \left[U(S \cup\{i\}) - U(S)\right]  
        \\& + \frac{1}{\binom{N-1}{|S|+1}} \left[U(S \cup\{j\}\cup\{i\}) -  U(S\cup\{j\})\right]\biggr\}\\
        = &~ \frac{1}{N}\sum\limits_{S \subseteq I \setminus\{i,j\}} \biggl\{\frac{1}{\binom{N-1}{|S|}} \left[U(S \cup\{j\}) - U(S)\right]  
        \\& + \frac{1}{\binom{N-1}{|S|+1}} \left[U(S \cup\{j\}\cup\{i\}) -  U(S\cup\{i\})\right]\biggr\}\\
        = &~\hat s_j.
    \end{align*}
    \end{small}
    \item \textbf{Zero element.} For any client $i \in I$. Suppose that for any subset of clients $S \subseteq I \setminus \{i\}$, $U(S \cup \{i\}) = U(S)$. Then, 
    \begin{small}
    \begin{align*}
        \hat s_i &= \frac{1}{N} \sum_{t=1}^T\sum\limits_{S \subseteq I \setminus\{i\}} \frac{1}{\binom{N-1}{|S|}} \left[ U_t(S \cup\{i\}) - U_t(S)\right] \\
        &= \frac{1}{N}\sum\limits_{S \subseteq I \setminus\{i\}} \frac{1}{\binom{N-1}{|S|}} \left[\sum_{t=1}^T U_t(S \cup\{i\}) - \sum_{t=1}^T U_t(S)\right]\\
        &= \frac{1}{N}\sum\limits_{S \subseteq I \setminus\{i\}} \frac{1}{\binom{N-1}{|S|}} \left[ U(S \cup\{i\}) - U(S)\right] = 0.
    \end{align*}
    \end{small}
    \item \textbf{Additivity.} If the utility function $U$ can be expressed as the sum of separate utility functions, namely $U = U_1 + U_2$ for some $U_1, U_2 : 2^I \to \mathbb{R}$, then for any client $i \in I$, we have
    \begin{small}
    \begin{align*}
        \hat s_i 
        = & \frac{1}{N}\sum\limits_{S \subseteq I \setminus\{i\}} \frac{1}{\binom{N-1}{|S|}} \left[ U(S \cup\{i\}) - U(S)\right] 
        \\= &\frac{1}{N}\sum\limits_{S \subseteq I \setminus\{i\}} \frac{1}{\binom{N-1}{|S|}} \bigg[ U_1(S \cup\{i\}) + U_2(S \cup\{i\}) 
        \\&- U_1(S) - U_2(S)\bigg] 
        \\=& \hat s_{1,i} + \hat s_{2, i},
    \end{align*}
    \end{small}
    where $\hat s_{1,i}$ and $\hat s_{2, i}$ are respectively the ComFedSVs computed from the true utility matrices $\mathcal{U}_1$ and $\mathcal{U}_2$.  
\end{itemize}

Next, we show that if $\|\mathcal{U} - WH^T\|_1 \leq \delta$, then $|s_i - \hat s_i| \leq \frac{2\delta}{N}$ for all $i$. By the definition of ComFedSV, we have 
\begin{small}
\begin{align*}
    &~|s_i - \hat s_i| 
    \\= &~\biggl| \frac{1}{N} \sum_{t=1}^T\sum\limits_{S \subseteq I \setminus\{i\}} \frac{1}{\binom{N-1}{|S|}} w_t^T(h_{S\cup\{i\}} - h_S) 
    \\&- \frac{1}{N} \sum_{t=1}^T\sum\limits_{S \subseteq I \setminus\{i\}} \frac{1}{\binom{N-1}{|S|}} \left[ U_t(S \cup\{i\}) - U_t(S)\right] \biggr|\\
    = &~\frac{1}{N}\biggl| \sum\limits_{S \subseteq I \setminus\{i\}} \frac{1}{\binom{N-1}{|S|}} \biggl[ \sum_{t=1}^T \left(w_t^Th_{S\cup\{i\}} - U_t(S\cup\{i\})\right) 
    \\&+  \sum_{t=1}^T \left(U_t(S) - w_t^Th_{S}\right) \biggr]\biggr|\\
    \leq &~\frac{1}{N}\sum\limits_{S \subseteq I \setminus\{i\}} \frac{1}{\binom{N-1}{|S|}}\biggl[ \sum_{t=1}^T \left|w_t^Th_{S\cup\{i\}} - U_t(S\cup\{i\})\right| 
    \\&+  \sum_{t=1}^T \left|U_t(S) - w_t^Th_{S}\right| \biggr]\\
    \leq &~\frac{1}{N}\sum\limits_{S \subseteq I \setminus\{i\}} \frac{2}{\binom{N-1}{|S|}} \|\mathcal{U} - WH^T\|_1 \\
    \leq &~\frac{2\delta}{N}.
\end{align*}
\end{small}
Similarly, we can show that $|s_{1,i} - \hat s_{1,i}| \leq \frac{2\delta_1}{N}$ and $|s_{2,i} - \hat s_{2,i}| \leq \frac{2\delta_2}{N}$. 

Finally, we show that $s_i$'s approximately satisfy the \textbf{Symmetry}, \textbf{Zero element} and \textbf{Additivity} properties within the tolerance of $\frac{4\delta}{N}$ as stated in Theorem 1.
\begin{itemize}
    \item \textbf{Symmetry.} 
    \[|s_i - s_j| \leq |\hat s_i - \hat s_j| + |\hat s_i - s_i| + |\hat s_j - s_j| \leq \frac{4\delta}{N}.\]
    \item \textbf{Zero element.}
    \[|s_i| \leq |\hat s_i| + |s_i - \hat s_i|\leq \frac{4\delta}{N}.\]
    \item \textbf{Additivity.}
    \begin{align*}
            &|s_i - (s_{1,i} + s_{2,i})| 
        \\ \leq &|\hat s_i - (\hat s_{1,i} + \hat s_{2,i})| + |s_i - \hat s_i| + |s_{1,i} - \hat s_{1,i}| + |s_{2,i} - \hat s_{2,i}|
        \\ \leq &\frac{2\delta}{N} + \frac{2\delta_1}{N} + \frac{2\delta_2}{N} 
        \\\leq &\frac{4\delta}{N}.
    \end{align*}
\end{itemize}
\end{proof}






\bibliographystyle{IEEEtran}
\bibliography{refs.bib}

\end{document}